\RequirePackage{fix-cm}
\documentclass[twocolumn]{svjour3}          
\smartqed  

\usepackage{cite}
\usepackage[pdftex]{graphicx}
\usepackage{ragged2e}
\usepackage[tight,footnotesize]{subfigure}
\usepackage{rotating}
\usepackage{graphicx}
\usepackage{amsmath,amssymb} 
\usepackage{array}
\usepackage{multirow}
\usepackage{colortbl}
\usepackage{amsfonts}
\usepackage{pifont}
\usepackage{xspace}
\usepackage{etoolbox}
\usepackage{overpic}
\usepackage{color}
\usepackage{microtype}

\usepackage{booktabs}
\usepackage{tabularray}
\usepackage{makecell}
\usepackage{enumitem}

\newcommand{\figref}[1]{Fig. \ref{#1}}
\newcommand{\tabref}[1]{Tab. \ref{#1}}

\newcommand{\orcid}[1]{\href{https://orcid.org/#1}{\includegraphics[width=10pt]{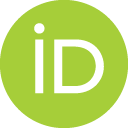}}}

\def\eg{\emph{e.g.}}
\def\ie{\emph{i.e.}}

\def\etal{{\em et al.}}

\newcommand{\tool}{RobustMQ\space}
\newcommand{\toolns}{RobustMQ}

\usepackage{hyperref}
\hypersetup{breaklinks=true,citecolor=blue, colorlinks}

\graphicspath{{./Imgs/}}
\DeclareGraphicsExtensions{.pdf,.jpg,.png}

\usepackage{silence}
\journalname{Research Article}

\begin{document}

\title{RobustMQ: Benchmarking Robustness of Quantized Models}

\titlerunning{RobustMQ}        

\author{Yisong Xiao \orcid{0000-0001-8227-0052}        
\and
  Aishan Liu \orcid{0000-0002-4224-1318} 
  \and 
  Tianyuan Zhang \orcid{0000-0001-9874-6828} \and 
  Haotong Qin \orcid{0000-0001-7391-7539} \and 
  Jinyang Guo \orcid{0000-0003-1956-3367} \and 
  Xianglong Liu \orcid{0000-0002-7618-3275} 
}

\authorrunning{Xiao \etal} 

\institute{
Yisong Xiao is with Shen Yuan Honors College and \emph{NLSDE}, Beihang University, Beijing, China. (Email: xiaoyisong@buaa.edu.cn). \\
Aishan Liu is with \emph{NLSDE}, Beihang University, Beijing, China, and Institute of Dataspace, Hefei, Anhui, China. (Email: liuaishan@buaa.edu.cn). \\
Tianyuan Zhang is with \emph{NLSDE}, Beihang University, Beijing, China. (Email: 19373397@buaa.edu.cn). \\
Haotong Qin is with ETH Zurich. (Email: qinhaotong@gmail.com). \\
Jinyang Guo is with Institute of Artificial Intelligence, and \emph{NLSDE}, Beihang University, Beijing, China. (Email: jinyangguo@buaa.edu.cn). \\
Xianglong Liu is with \emph{NLSDE}, Beihang University, Beijing, China, Zhongguancun Laboratory, Beijing, China, and Institute of Dataspace, Hefei, Anhui, China. (Email: xlliu@buaa.edu.cn). \\
Corresponding author: Aishan Liu and Xianglong Liu.
}

\date{Received: date / Accepted: date}

\maketitle

\begin{abstract}

Quantization has emerged as an essential technique for deploying deep neural networks (DNNs) on devices with limited resources. However, quantized models exhibit vulnerabilities when exposed to various noises in real-world applications. Despite the importance of evaluating the impact of quantization on robustness, existing research on this topic is limited and often disregards established principles of robustness evaluation, resulting in incomplete and inconclusive findings. To address this gap, we thoroughly evaluated the robustness of quantized models against various noises (adversarial attacks, natural corruptions, and systematic noises) on ImageNet. 
The comprehensive evaluation results empirically provide valuable insights into the robustness of quantized models in various scenarios, for example: 
(1) quantized models exhibit higher adversarial robustness than their floating-point counterparts, but are more vulnerable to natural corruptions and systematic noises; (2) in general, increasing the quantization bit-width results in a decrease in adversarial robustness, an increase in natural robustness, and an increase in systematic robustness; (3) among corruption methods, \textit{impulse noise} and \textit{glass blur} are the most harmful to quantized models, while \textit{brightness} has the least impact; (4) among systematic noises, the \textit{nearest neighbor interpolation} has the highest impact,  while bilinear interpolation, cubic interpolation, and area interpolation are the three least harmful. Our research contributes to advancing the robust quantization of models and their deployment in real-world scenarios.


\end{abstract}

\section{Introduction}

\label{sec:intro}

Deep neural networks (DNNs) have demonstrated impressive performance in a broad range of applications, including computer vision \cite{krizhevsky2017imagenet,DBLP:conf/cvpr/ZhaoZXLP22}, natural language processing \cite{bahdanau2014neural, sutskever2014sequence}, and speech recognition \cite{hinton2012deep, graves2013speech}. However, the deployment of large DNNs on resource-constrained devices, such as smartphones and embedded systems, poses challenges due to their high memory and computational requirements. To address this issue, researchers have proposed various model compression techniques, including model quantization \cite{Qin_2020_CVPR,Zhang_2021_CVPR,li2021mqbench,qin2023bibench,qin2022bibert}, pruning \cite{guo2021jointpruning,guo2020multi,guo2020channel,guo2020model,guo2023multidimensional}, and neural network distillation \cite{bucilu2006model, hinton2015distilling}. Among these techniques, model quantization has become a critical approach for compressing DNNs due to its ability to maintain network structure and achieve comparable performance simultaneously. By mapping the network parameters from continuous 32-bit floating-point (FP) numbers to discrete low-bit integers, model quantization achieves reduced memory usage and faster inference, making it well-suited for resource-constrained devices. 

While model quantization offers a viable solution for deploying DNNs on resource-constrained devices, it also presents challenges in ensuring the trustworthiness (such as robustness, fairness, privacy, and etc) of quantized models in the real world \cite{guo2023comprehensive,liu2019perceptual,liu2020bias,liu2020spatiotemporal,liu2021ANP,liu2021training,liu2022harnessing,liu2023x,wang2021dual,zhang2021interpreting,xiao2023latent}. 
DNNs are highly susceptible to adversarial examples \cite{liu2019perceptual,goodfellow2014explaining,liu2023x,liu2022harnessing}, which are perturbations carefully designed to be undetectable to human vision but can easily deceive DNNs, posing a significant threat to practical deep learning applications. For example, the placement of three small adversarial stickers on a road intersection can cause the Tesla Autopilot system \cite{boloor2020attacking} to misinterpret the lane markings and swerve into the wrong lane, which severely risks people's lives and even lead to potential death. In addition, DNNs are vulnerable to natural corruptions \cite{hendrycks2019benchmarking} such as snow and motion blur, which are common in real-world scenarios and can significantly reduce the accuracy of DNN models. Moreover, system noises resulting from the mismatch between software and hardware can also have a detrimental impact on model accuracy \cite{wang2021real}. These vulnerabilities demonstrate that quantized networks deployed in safety-critical applications (like autonomous driving and face recognition) are unreliable when faced with various perturbations in real-world scenarios.

Therefore, it is critical to conduct a comprehensive evaluation of the robustness of quantized models before their deployment to identify potential weaknesses and unintended behaviors. In recent years, researchers have developed robustness benchmarks \cite{tang2021robustart,croce2020robustbench,wang2021adversarial,yi2021benchmarking,zhang2023benchmarking} tailored for assessing the robustness of deep learning models, employing multiple adversarial attack methods to thoroughly evaluate floating-point networks across various tasks. Through extensive experiments, researchers have revealed and substantiated several insights, such as the observation that larger model parameter sizes often lead to better adversarial robustness \cite{tang2021robustart,madry2017towards}, which highlights the significance of model complexity in determining robustness. While numerous studies have extensively investigated the robustness of floating-point networks, research on the robustness of quantized models \cite{bernhard2019impact,lin2019defensive,alizadeh2020gradient,xiao2023benchmarking} remains inadequate, lacking diversity in terms of noise sources and relying solely on small datasets. Consequently, the existing literature fails to thoroughly assess the robustness of quantized models, leading to a gap in the understanding of their vulnerabilities and strengths.

To bridge this gap, we build \toolns, a comprehensive robustness evaluation benchmark for quantized models. Our benchmark systematically assesses the robustness of quantized models using 3 popular quantization methods (\ie, DoReFa \cite{zhou2016dorefa}, PACT \cite{choi2018pact}, and LSQ \cite{esser2019learned}) and 4 classical architectures (ResNet18 \cite{he2016deep}, ResNet50 \cite{he2016deep}, RegNetX600M \cite{radosavovic2020designing}, and MobileNetV2 \cite{sandler2018mobilenetv2}). Each method is evaluated for four commonly used bit-widths. To thoroughly study the robustness of quantized models against noises originating from different sources, our analysis comprises 3 progressive adversarial attacks (covering $\ell_{1},\ell_{2},\ell_{\infty}$ magnitudes, along with three different perturbation budgets), 15 natural corruptions, and 14 systematic noises on the ImageNet benchmark. Our empirical results demonstrate that lower-bit quantized models exhibit better adversarial robustness but are more susceptible to natural corruptions and systematic noises. In summary, increasing the quantization bit-width generally leads to a decrease in adversarial robustness, an increase in natural robustness, and an increase in systematic robustness. Moreover, our findings indicate that \textit{impulse noise} and \textit{glass blur} are the most harmful corruption methods for quantized models, while \textit{brightness} has the least impact. Additionally, among systematic noises, the \textit{nearest neighbor interpolation} has the highest impact, while bilinear interpolation, cubic interpolation, and area interpolation are the three least harmful. Our main contributions can be summarized as follows:

\begin{itemize}
    \item To the best of our knowledge, \tool is the first to comprehensively evaluate the robustness of quantized models.  \tool covers three popular quantization methods, four common bit-widths, and four classical architectures across a range of noise types, including adversarial attacks, natural corruptions, and systematic noises.
    \item Through extensive experiments, \tool uncovers valuable insights into the robustness of quantized models, shedding light on their strengths and weaknesses in comparison to floating-point models across various scenarios.
    \item The RobustMQ benchmark provides a standardized framework for evaluating the robustness of quantized models, enabling further research and development in this field. It is publicly available on our website \cite{ourweb}.
\end{itemize}

\section{Related Work}
\label{sec:related_work}
In this section, we review related works on network quantization, adversarial attacks, and the recent advancements in the integration of these fields.

\subsection{Network quantization} 

Network quantization compresses DNN models by reducing the number of bits required to represent each weight to save memory usage and speed up hardware inference \cite{gholami2021survey}. A classic quantization process involves both \textit{quantization} and \textit{de-quantization} operations. The quantization function maps real values $r$ to integers, while the de-quantization function allows approximate recovery of real values $r^{'}$ from the quantized values. This process can be mathematically formulated as:
\begin{equation}
Q(r)=\mathbf{Int}(r/S)-Z,\quad r^{'}=S \cdot (Q(r)+Z)
\end{equation}
where $Q$ is the quantization operator, $r$ and $r^{'}$ is real value and de-quantized real value respectively, $S$ and $Z$ denote \textit{scale} and \textit{zero-point} respectively. Given $t$ bits, the range after quantization is determined by $[-2^{t-1},2^{t-1}-1]$. After quantization, the recovered real values $r^{'}$ may not exactly be equal to the original values $r$ due to the rounding operation.

Quantization methods can be broadly classified into two strategies: Post-Training Quantization (PTQ) and Quantization-Aware Training (QAT). PTQ methods are applied after the model is fully trained, without any adjustments to the model during the training process, which often results in lower accuracy. On the other hand, QAT methods involve fine-tuning or retraining the model with training data to achieve higher accuracy in the quantized form. Thus, we primarily focus on QAT methods, and we here provide a brief review of the commonly used QAT methods. 

One string of research designed rules to fit the quantizer to the data distribution \cite{li2016ternary,zhou2016dorefa}. For example, DoReFa-Net \cite{zhou2016dorefa} simply clips the activation to $[0,1]$ and then quantizes it, due to the observation that most activation falls into this range in many network architectures (\eg, AlexNet \cite{krizhevsky2017imagenet} and ResNet \cite{he2016deep}). Other notable work focused on learning appropriate quantization parameters during the backpropagation process \cite{esser2019learned,jung2019learning,choi2018pact}.
PACT \cite{choi2018pact} clip the activation by a handcrafted parameter and optimize the clipping threshold. However, it is important to note that PACT has no gradient below the clip point, which can lead to gradient vanishing problems during backpropagation. Notice the limitation in PACT, LSQ \cite{esser2019learned} estimates the gradient at each weight and activation layer to adaptively adjust the step size of quantization. By learning the \textit{scale} alongside network parameters, LSQ is able to achieve more fine-grained quantization, which improves the accuracy of quantized models.

\subsection{Adversarial attacks} 

Adversarial examples are inputs with small perturbations that could easily mislead the DNNs \cite{goodfellow2014explaining}. 
Formally, given a DNN $f_{\Theta}$ and an input $\mathbf{x}$ with the ground truth label $\mathbf{y}$, an adversarial example $\mathbf{x}_{adv}$ satisfies
\begin{equation}
f_{\Theta}(\mathbf{x}_{adv}) \neq \mathbf{y} \quad s.t. \quad \|\mathbf{x}-\mathbf{x}_{adv}\| \leq \epsilon,
\end{equation}
where $\|\cdot\|$ is a distance metric and commonly measured by the $\ell_{p}$-norm ($p\in$\{1,2,$\infty$\}).

A long line of work has been dedicated to performing adversarial attacks \cite{goodfellow2014explaining,madry2017towards,croce2020reliable,liu2020bias,liu2019perceptual,liu2020spatiotemporal,wang2021dual}, which can be mainly divided into white-box and black-box manners based on access to the target model. For white-box attacks, adversaries have complete knowledge of the target model and can fully access it; while for black-box attacks, adversaries have limited or even without any knowledge of the target model and can not directly access it. This paper primarily employs white-box attacks to evaluate the adversarial robustness of target models, as they offer stronger attack capabilities. Here, we introduce the attack methods relevant to our benchmark. 

\textbf{Fast Gradient Sign Method (FGSM)}. FGSM \cite{goodfellow2014explaining} is a one-step attack method with $\ell_{\infty}$-norm. It calculates the gradient of the loss function with respect to the input only once and subsequently adds gradient noise to generate an adversarial example. Although FGSM has a relatively weaker attack capability, it is computationally efficient in generating adversarial examples.

\textbf{Projected Gradient Descent (PGD)}. PGD \cite{madry2017towards} is regarded as one of the most powerful and widely used attacks due to its high attack success rate. It builds upon the FGSM by introducing an iterative process with a gradient projecting at each step. 

\textbf{AutoAttack}. Croce \etal \cite{croce2020reliable} proposed two automatic step size adjustment methods (APGD-CE and APGD-DLR) to address problems such as suboptimal step size in PGD. Then they combined two existing complementary attacks to form a parameter-free and computationally affordable attack (\ie, AutoAttack). AutoAttack has demonstrated superior performance compared to state-of-the-art attacks, thus becoming a crucial tool for evaluating model robustness.

\subsection{Robustness of quantized models}  

A number of studies have been proposed to evaluate the robustness of floating-point networks \cite{croce2020robustbench,zhang2021interpreting,tang2021robustart,liu2021training,wang2022defensive}. 
For instance, Croce \etal \cite{croce2020robustbench} introduced RobustBench, a benchmark based on the CIFAR dataset, which employs AutoAttack to assess the robustness of models strengthened by various defense methods, including adversarial training, gradient masking, and label smoothing. In contrast to the evaluation of defense methods, Tang \etal \cite{tang2021robustart} focused on investigating the robustness of model structures and training techniques using the large-scale ImageNet dataset, offering valuable insights for training robust models. 

Comparatively, Merkle \etal \cite{merkle2022pruning} benchmarked the adversarial robustness of pruned models for several pruning methods, revealing that pruned models enjoy better robustness against adversaries. 
However, the robustness of quantized networks has been relatively underexplored.  
Bernhard \etal \cite{bernhard2019impact} utilized an ensemble of quantized models to filter adversarial examples, given that current adversarial attacks demonstrate limited transferability against quantized models. Lin \etal \cite{lin2019defensive} proposed a defensive quantization method to suppress the amplification of adversarial noise during propagation by controlling the Lipschitz constant of the network during quantization. Similarly, Alizadeh \etal \cite{alizadeh2020gradient} also designed a regularization scheme to improve the robustness of the quantized model by controlling the magnitude of adversarial gradients. In this paper, we aim to thoroughly evaluate the robustness of quantized models against multiple noises for several quantization methods, architectures, and quantization bits.

\section{RobustMQ Benchmark}
\label{sec:benchmark}

Existing research on the impact of quantization compression on neural network robustness is fragmented and lacks adherence to established principles in robustness evaluation. To address this issue, this paper proposes \toolns, a comprehensive robustness evaluation benchmark for quantized models with consistent settings. \tool provides researchers with a valuable tool to gain deeper insights into the impact of various perturbations on quantized model robustness, aiding in the development of more robust quantization methods for deploying reliable and secure deep learning models in real-world scenarios. The \tool benchmark encompasses adversarial robustness, natural robustness, and systematic robustness, considering three quantization methods, four bit-widths, four architectures, three progressive adversarial attack methods (covering three magnitudes and three perturbation budgets), fifteen natural corruptions, and fourteen systematic noises on ImageNet.


\subsection{Robustness evaluation approaches} 

Quantized models that are extensively deployed in edge devices are vulnerable to various perturbations in real-world scenarios. In accordance with the guidelines proposed by Tang \etal \cite{tang2021robustart}, we classify these perturbations into adversarial attacks, natural corruptions, and systematic noises, and leverage them to thoroughly evaluate the robustness of quantized models. 

\subsubsection{Adversarial attacks}
To model the worst-case scenario (\ie, strongest adversaries), we consider attacks conducted under a white-box manner where the adversary has full access to the model architecture, training data, and gradient information. Specifically, we employ FGSM-$\ell_{\infty}$, PGD-$\ell_{1}$, PGD-$\ell_{2}$, PGD-$\ell_{\infty}$ and AutoAttack-$\ell_{\infty}$ to craft adversarial perturbations. These three attack methods form a progressive evaluation, where their computing resource consumption and attack capabilities are improved, enabling a comprehensive assessment of the quantized model's adversarial robustness. Furthermore, we set three progressive perturbation budgets (small, middle, and large) for each attack method.  

\subsubsection{Natural corruptions}
To simulate natural corruptions, we utilize 15 distinct perturbation methods from ImageNet-C benchmark \cite{hendrycks2019benchmarking}. These methods can be categorized into four groups: (1) noise, which includes \textit{gaussian noise}, \textit{shot noise}, and \textit{impulse noise}; (2) blur, which includes \textit{defocus blur}, \textit{frosted glass blur}, \textit{motion blur}, and \textit{zoom blur}; (3) weather, which includes \textit{snow}, \textit{frost}, \textit{fog}, and \textit{brightness}; and (4) digital, which includes \textit{contrast}, \textit{elastic}, \textit{pixelation}, and \textit{JPEG compression}. Each corruption type is evaluated at five levels of severity to account for variations in the intensity of corruptions. Thus, we have 75 perturbation methods in total for natural corruptions evaluation. In addition to single corruption images, we also investigate corruption sequences generated from ImageNet-P \cite{hendrycks2019benchmarking}. Each sequence in ImageNet-P comprises more than 30 frames, allowing us to study the robustness of quantized models against dynamic and continuous corruptions.

\subsubsection{Systematic noises}
Moreover, system noises are always present when models are deployed in edge devices due to changes in hardware or software. To assess the influence of system noises on quantized models, we incorporate pre-processing operations from ImageNet-S \cite{wang2021real}, which involve image decoding and image resize processes. Image decoding refers to the process of converting an original image file into an RGB channel map tensor, where the inverse discrete cosine transform (iDCT) serves as a core step. However, discrepancies in the implementation of iDCT among various image processing libraries result in variations in the output. As a consequence, the pixel values of the final RGB tensor are affected, leading to slight differences in the decoded images. Therefore, we employ the decoders from third-party libraries such as Pillow \cite{umesh2012image}, OpenCV \cite{bradski2000opencv}, and FFmpeg \cite{tomar2006converting} to obtain systematic noises. Additionally, image resize is utilized to change the image resolution. In the resize process, different interpolation algorithms are employed to predict the new pixel positions, potentially leading to slight variations in the new pixel values. Thus, for different image resize methods, we incorporate bilinear, nearest, cubic, hamming, lanczos, area, and box interpolation modes from the OpenCV and Pillow tools. In total, systematic noises consist of three frequently used decoders and seven commonly used resize modes.

\subsection{Evaluation metrics} 

\subsubsection{Adversarial robustness} 
For specific adversarial attacks, we adopt the model accuracy to measure adversarial robustness ($AR$), which is calculated by subtracting the Attack Success Rate (ASR) from 1 (\ie, \textit{$1-ASR$}). Mathematically, $AR$ can be calculated with the following expression:
\begin{equation}
AR = 1 - P_{(\mathbf{x},\mathbf{y})\sim \mathcal{D}}{(f(\mathcal{A}_{\epsilon,p}^{f}(\mathbf{x})) \neq \mathbf{y}) }, 
\end{equation}
where $f$ is the target tested model, $\mathcal{D}$ is the validation set, $\mathcal{A}_{\epsilon,p}$ represents the adversary, $\epsilon$ and $p$ denotes the perturbation budget and distance norm respectively. This metric quantifies the model's capability to retain normal functioning under attacks, with higher $AR$ indicating a stronger model against the specific adversarial attack. While we aim to evaluate the robustness among models with different clean accuracy, it is crucial to measure the relative performance drop against adversarial attacks, denoted as $AAI$ (Adversarial Attack Impact): 
\begin{equation}
AAI = \frac{ACC - AR}{ACC}, 
\end{equation}
where $ACC$ represents the clean accuracy. A lower $AAI$ value indicates that the models are more robust.

For the union of different attacks, we adopt Worst-Case Adversarial Robustness ($WCAR$) to measure adversarial robustness (a higher value indicates a more robust model) against them:
\begin{equation}
WCAR = 1 - P_{(\mathbf{x},\mathbf{y})\sim \mathcal{D}}{\mathrm{Any}_{\mathcal{A} \in \mathcal{A}s}(f(\mathcal{A}_{\epsilon,p}^{f}(\mathbf{x})) \neq \mathbf{y}) }, 
\end{equation}
where $\mathcal{A}s$ represents a set of adversaries, $\mathrm{Any}(\cdot)$ is a function that returns true if any of the adversary $\mathcal{A}$ in  $\mathcal{A}s$ attacks successfully. $WCAR$ represents a lower bound of model adversarial robustness against various adversarial attacks.  Specifically, We further refine $WCAR$ based on the perturbation budget employed in adversarial attacks: $WCAR \ (\text{small} \ \epsilon)$, $WCAR \ (\text{middle} \ \epsilon)$, and $WCAR \ (\text{large} \ \epsilon)$.

\subsubsection{Natural robustness} 
Natural robustness measures the accuracy of a model in maintaining its performance after being perturbed by natural noise. Therefore, given a corruption method $c$, we calculate the accuracy as its natural robustness:
\begin{equation}
ACC_{c}=P_{(\mathbf{x},\mathbf{y})\sim \mathcal{D}}{(f(c(\mathbf{x}))=\mathbf{y})}.
\end{equation}
Similar to $AAI$ in adversarial robustness, we also define $NCI$ (Natural Corruption Impact) to measure the relative performance drop against natural corruptions:
\begin{equation}
NCI = \frac{ACC - ACC_{c}}{ACC}.
\end{equation}

To aggregate the evaluation results among 15 corruptions, We adopt the average accuracy of the quantized model on all corruptions to measure the mean natural robustness, denoted as $mNR$:
\begin{equation}
mNR = \mathbb{E}_{c\sim C}({P_{(\mathbf{x},\mathbf{y})\sim \mathcal{D}}{(f(c(\mathbf{x}))=\mathbf{y})}}),
\end{equation}
where $C$ denotes the set of corruption methods. A higher value of $mNR$ means better natural robustness. The average relative performance drop can be calculated by $(ACC-mNR) / ACC$.

As for the corruption sequence $\mathcal{S}$, we utilize ``Flip Probability'' of model predictions to measure its robustness, denoted as $FP$:
\begin{equation}
FP = P_{\mathbf{x}\sim \mathcal{S}}{(f(\mathbf{x}_{j}) \neq f(\mathbf{x}_{j-1}))}.
\end{equation}

For sequences generated by multiple corruption methods, we average their Flip Probability to obtain $mFP$ (\ie, mean Flip Probability). Note that a lower $FP$ value indicates a model that performs more stably in the presence of dynamic and continuous corruptions.

\subsubsection{Systematic robustness}
Systematic robustness measures a model's ability to withstand various software-dependent and component-dependent system noise attacks in diverse deployment environments. For a given decode or resize method $s$, we compute the accuracy as its systematic robustness:
\begin{equation}
ACC_{s}=P_{(\mathbf{x},\mathbf{y})\sim \mathcal{D}}{(f(s(\mathbf{x}))=\mathbf{y})}.
\end{equation}
To emphasize the impact of noise, we introduce $SNI$ (Systematic Noise Impact) as a metric to quantify the relative performance drop:
\begin{equation}
SNI = \frac{ACC - ACC_{s}}{ACC}.
\end{equation}

Furthermore, to evaluate the robustness among all systematic noises, we calculate the standard deviation of their $ACC_{s}$ as systematic robustness ($SR$):
\begin{equation}
SR = \mathbb{D}_{s\sim S}({P_{(\mathbf{x},\mathbf{y})\sim \mathcal{D}}{(f(s(\mathbf{x}))=\mathbf{y})}}),
\end{equation}
where $S$ denotes a set of decode or resize methods. A lower value of $SR$ means better stability towards different systematic noises.

\subsection{Evaluation objects}

\subsubsection{Dataset} 
Our \tool aims to obtain broadly applicable results and conclusions for quantized models in the computer vision field. Therefore, we focus on the basic image classification tasks and employ the large-scale ImageNet \cite{deng2009imagenet} dataset. In contrast to widely used small-scale datasets like MNIST \cite{lecun1998mnist}, CIFAR-10 \cite{krizhevsky2009learning}, and CIFAR-100 \cite{krizhevsky2009learning}, ImageNet provides a more extensive collection of images and classes, making it more suitable as a benchmark for testing models in robustness evaluation. The ImageNet dataset comprises 1.2 million training images and 50,000 validation images, covering a total of 1,000 different classes.

\subsubsection{Network architectures}
Our \tool contains four architectures, including ResNet18\cite{he2016deep}, ResNet50\cite{he2016deep}, RegNetX600M \cite{radosavovic2020designing}, and MobileNetV2 \cite{sandler2018mobilenetv2}. 
(1) ResNet18 and ResNet50 are classical backbone architectures that have proven to be highly effective in various computer vision tasks. Both architectures are built on the concept of residual blocks, which employ skip connections to mitigate the vanishing gradient problem and facilitate the training of deep networks. 
(2) RegNetX600M is an advanced architecture discovered through model structure search, specifically optimized to achieve efficient and powerful feature extraction. It leverages group convolution to enable parallel processing and significantly reduce computational complexity, making it ideal for resource-constrained edge devices.
(3) MobileNetV2 is a lightweight network designed for efficient deployment on edge devices. It employs depthwise separable convolutions, which separate the spatial and channel-wise convolutions, reducing the computational burden while maintaining performance.

\subsubsection{Quantization methods}
Within \toolns, we concentrate on three widely used quantization methods: DoReFa \cite{zhou2016dorefa}, PACT \cite{choi2018pact}, and LSQ \cite{esser2019learned}. LSQ and DoReFa methods perform their quantization methods both on model weights and activation values. On the other hand, PACT applies its specific parameter truncation to quantize activation values, while leveraging the method in DoReFa for weight quantization. For the choice of quantization bits, we adopt the commonly used set in deployments (\ie, 2, 4, 6, and 8). For each architecture, we quantize models on the ImageNet training set starting from the same floating-point model, then evaluate their robustness against perturbations generated on the ImageNet validation set.


\section{Experiments and Analysis}
In this section, we showcase the evaluation results of quantized models under different noises. Subsequently, we consolidate several conclusive insights gleaned from the evaluations by addressing the following research questions: (1) How robust are quantized models compared with FP models? (2) Which quantization method or bit-width exhibits greater robustness against perturbations?  (3) What is the impact of architecture or size on the robustness of quantized models? (4) To which type of noise is the quantized model most susceptible?


\subsection{Clean accuracy}


\tabref{tab:clean_acc} reports the clean accuracies of quantized models. Most of the quantized models maintain comparable accuracy to their 32-bit pre-trained models, while certain quantization methods may struggle to maintain comparable accuracy when using low bit-widths (\eg, 2-bit). Among the quantized models evaluated, 12 models fails to converge. For example, ResNet18 PACT 2-bit achieves a mere 2.61\% accuracy. Therefore, we label these models as ``NC" and  exclude them from our evaluations to ensure fair and reliable assessments of robustness.

\begin{table}[ht] \small
\centering

\caption{Clean accuracy of quantized models and FP models. ``NC" denotes not converged.}

\label{tab:clean_acc}

\resizebox{\columnwidth}{!}{%
\begin{tabular}{@{}llccccc@{}}
\toprule
Model & Method & 2-bit & 4-bit & 6-bit & 8-bit & 32-bit \\ \midrule
\multirow{3}{*}{ResNet18} & DoReFa & 62.31 & 70.60 & 71.15 & \pmb{71.51} & \multirow{3}{*}{71.06} \\
 & PACT & NC & 70.41 & 71.17 & 71.43 &  \\
 & LSQ & 65.97 & 70.52 & 71.10 & 71.31 &  \\ \midrule
\multirow{3}{*}{ResNet50} & DoReFa & NC & 76.81 & 77.11 & 77.01 & \multirow{3}{*}{76.63} \\
 & PACT & NC & 76.80 & 77.02 & 77.06 &  \\
 & LSQ & 69.83 & 76.99 & 77.36 & \pmb{77.54} &  \\ \midrule
\multirow{3}{*}{RegNetX600M} & DoReFa & NC & 72.70 & 73.79 & 74.00 & \multirow{3}{*}{73.55} \\
 & PACT & NC & 72.16 & 73.80 & \pmb{74.05} &  \\
 & LSQ & NC & 72.54 & 73.64 & 73.77 &  \\ \midrule
\multirow{3}{*}{MobileNetV2} & DoReFa & NC & NC & NC & NC & \multirow{3}{*}{72.62} \\
 & PACT & NC & 70.41 & 72.52 & \pmb{72.70} &  \\
 & LSQ & NC & 70.32 & 72.05 & 72.40 &  \\ \bottomrule
\end{tabular}
}
\end{table}

\begin{table}[t] 
\centering

\caption{Worst-Case Adversarial Robustness of quantized models under all adversarial attacks with small budgets. Results are shown in $WCAR \ (small \ \epsilon)\textcolor{red}{\uparrow}$.}

\label{tab:ADV_WCAR}

\resizebox{\columnwidth}{!}{%
\begin{tabular}{@{}llccccc@{}}
\toprule
Model & Method & 2-bit & 4-bit & 6-bit & 8-bit & 32-bit \\ \midrule
\multirow{3}{*}{ResNet18} & DoReFa & \pmb{5.55} & 2.86 & 1.11 & 1.49 & \multirow{3}{*}{1.30} \\
 & PACT & NC & 4.03 & 1.94 & 1.50 &  \\
 & LSQ & 3.73 & 1.78 & 1.49 & 1.36 &  \\ \midrule
\multirow{3}{*}{ResNet50} & DoReFa & NC & 12.66 & 5.83 & 6.12 & \multirow{3}{*}{3.01} \\
 & PACT & NC & \pmb{13.71} & 7.78 & 6.21 &  \\
 & LSQ & 11.12 & 7.22 & 4.38 & 3.48 &  \\ \midrule
\multirow{3}{*}{RegNetX600M} & DoReFa & NC & 2.74 & 0.98 & 1.34 & \multirow{3}{*}{0.83} \\
 & PACT & NC & \pmb{6.21} & 1.91 & 1.37 &  \\
 & LSQ & NC & 2.05 & 1.29 & 1.11 &  \\ \midrule
\multirow{3}{*}{MobileNetV2} & DoReFa & NC & NC & NC & NC & \multirow{3}{*}{0.32} \\
 & PACT & NC & \pmb{6.73} & 1.03 & 0.79 &  \\
 & LSQ & NC & 5.39 & 2.31 & 1.28 &  \\ \bottomrule
\end{tabular}
}

\end{table}

\subsection{Evaluation of adversarial attacks}

\begin{table*}[!ht] 
\centering
\caption{Adversarial Robustness of models under FGSM-$\ell_{\infty}$ attack with small budget ($\epsilon=0.5/255$). Results are shown in $AR\textcolor{red}{\uparrow}(AAI\textcolor{green}{\downarrow})$.}
\label{tab:ADV_AR_FGSMLinf}
\begin{tabular}{@{}llrrrrr@{}}
\toprule
Model & Method & 2-bit & 4-bit & 6-bit & 8-bit & 32-bit \\ \midrule
\multirow{3}{*}{ResNet18} & DoReFa & 37.66 (39.56\%) & 31.50 (55.38\%) & 26.06 (63.37\%) & 25.79 (63.94\%) & \multirow{3}{*}{26.36 (62.90\%)} \\ 
 & PACT & NC & 36.49 (48.17\%) & 29.19 (58.99\%) & 25.66 (64.08\%) \\ 
 & LSQ & \pmb{46.85 (28.98\%)} & 35.95 (49.02\%) & 32.24 (54.66\%) & 31.07 (56.43\%) \\  \midrule
\multirow{3}{*}{ResNet50} & DoReFa & NC & 56.51 (26.43\%) & 48.56 (37.03\%) & 47.39 (38.46\%) & \multirow{3}{*}{45.57 (40.53\%)} \\ 
 & PACT & NC & 57.76 (24.79\%) & 50.28 (34.72\%) & 47.20 (38.75\%) \\ 
 & LSQ & \pmb{60.08 (13.96\%)} & 55.20 (28.30\%) & 50.77 (34.37\%) & 49.02 (36.78\%) \\  \midrule
\multirow{3}{*}{RegNetX600M} & DoReFa & NC & 44.98 (38.13\%) & 38.75 (47.49\%) & 36.80 (50.27\%) & \multirow{3}{*}{35.30 (52.01\%)} \\ 
 & PACT & NC & \pmb{51.45 (28.70\%)} & 41.09 (44.32\%) & 37.05 (49.97\%) \\ 
 & LSQ & NC & 48.31 (33.40\%) & 43.18 (41.36\%) & 40.67 (44.87\%) \\  \midrule
\multirow{3}{*}{MobileNetV2} & DoReFa & NC & NC & NC & NC & \multirow{3}{*}{24.48 (66.29\%)} \\ 
 & PACT & NC & 48.41 (31.25\%) & 29.51 (59.31\%) & 26.86 (63.05\%) \\ 
 & LSQ & NC & \pmb{53.38 (24.09\%)} & 41.19 (42.83\%) & 35.88 (50.44\%) \\  \bottomrule
 \end{tabular}
\end{table*}

\begin{table*}[!ht] 
\centering
\caption{Adversarial Robustness of quantized models under PGD-$\ell_{\infty}$ attack with small budget ($\epsilon=0.5/255$). Results are shown in $AR\textcolor{red}{\uparrow}(AAI\textcolor{green}{\downarrow})$. }

\label{tab:ADV_AR_PGDLinf}
\begin{tabular}{@{}llccccc@{}}
\toprule
Model & Method & 2-bit & 4-bit & 6-bit & 8-bit & 32-bit \\ \midrule
\multirow{3}{*}{ResNet18} & DoReFa & 26.61 (57.29\%) & 19.39 (72.54\%) & 13.57 (80.93\%) & 13.03 (81.78\%) & \multirow{3}{*}{13.96 (80.35\%)} \\ 
 & PACT & NC & 25.06 (64.41\%) & 17.12 (75.94\%) & 13.09 (81.67\%) \\ 
 & LSQ & \pmb{33.01 (49.96\%)} & 21.27 (69.84\%) & 17.54 (75.33\%) & 16.21 (77.27\%) \\  \midrule
\multirow{3}{*}{ResNet50} & DoReFa & NC & 42.07 (45.23\%) & 30.25 (60.77\%) & 28.04 (63.59\%) & \multirow{3}{*}{25.28 (67.01\%)} \\ 
 & PACT & NC & 44.30 (42.32\%) & 33.10 (57.02\%) & 27.84 (63.87\%) \\ 
 & LSQ & \pmb{48.25 (30.90\%)} & 37.78 (50.93\%) & 30.90 (60.06\%) & 27.99 (63.90\%) \\  \midrule
\multirow{3}{*}{RegNetX600M} & DoReFa & NC & 26.27 (63.87\%) & 17.38 (76.45\%) & 14.85 (79.93\%) & \multirow{3}{*}{13.15 (82.12\%)} \\ 
 & PACT & NC & \pmb{34.82 (51.75\%)} & 20.91 (71.67\%) & 14.85 (79.95\%) \\ 
 & LSQ & NC & 27.89 (61.55\%) & 20.83 (71.71\%) & 18.48 (74.95\%) \\  \midrule
\multirow{3}{*}{MobileNetV2} & DoReFa & NC & NC & NC & NC & \multirow{3}{*}{8.80 (87.88\%)} \\ 
 & PACT & NC & 36.06 (48.79\%) & 15.68 (78.38\%) & 12.29 (83.09\%) \\ 
 & LSQ & NC & \pmb{38.84 (44.77\%)} & 26.46 (63.28\%) & 20.41 (71.81\%) \\  \bottomrule

\end{tabular}
\end{table*}

\begin{figure*}[ht]
  \includegraphics[width=1.0\linewidth]{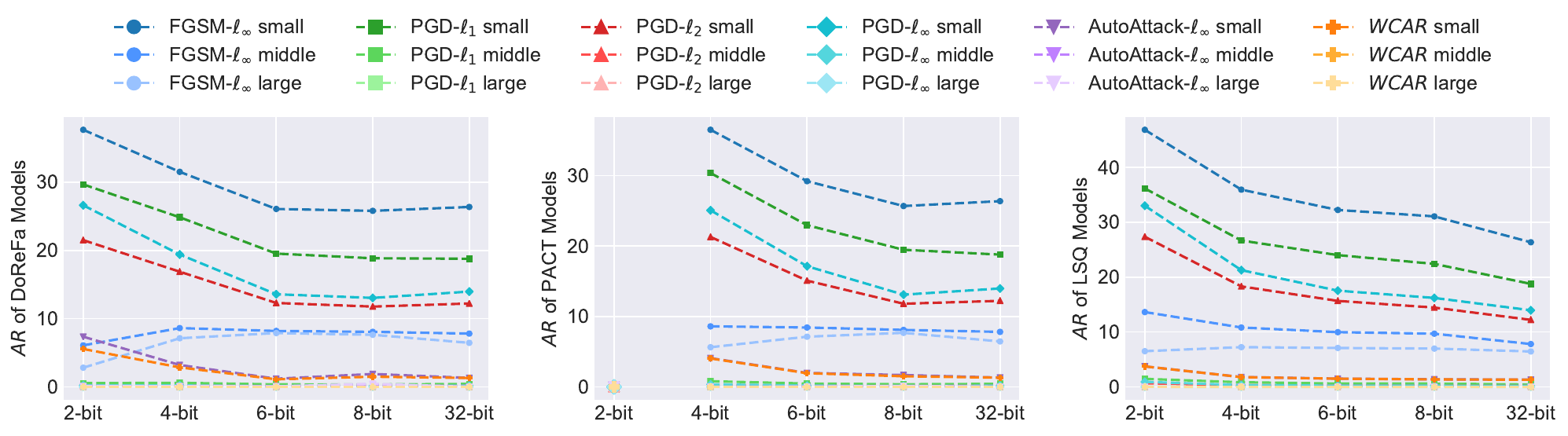}
  \caption{Adversarial robustness of ResNet18 models under specific attacks. From left to right: quantized by DoReFa, PACT, and LSQ respectively. The ``NC" models are assigned a $AR$ value of 0 in the figure. Similar trends, with $AR$ decreasing as the bit-width increases, are observed in the results of other architectures as well.}
  \label{fig:ResNet18_Adv}       
\end{figure*}

Under medium and high perturbation budgets, the values of $AR$ and $WCAR$ degrade significantly to almost 0 due to the increasing attack abilities. Therefore, in this section, we primarily present the results under small budgets to highlight the differences between different models. 
The robustness evaluation results with small budgets under all adversarial attacks, including FGSM-$\ell_{\infty}$ and PGD-$\ell_{\infty}$ attacks, are shown in \tabref{tab:ADV_WCAR}, \tabref{tab:ADV_AR_FGSMLinf}, and \tabref{tab:ADV_AR_PGDLinf}, respectively. And other adversarial robustness evaluation results can be found on our website \cite{ourweb}. From the results, we could make several observations for quantized models as follows.

(1)\emph{Better adversarial robustness vs FP models}. Unlike the decrease in clean accuracy, quantized models exhibit higher worst-case adversarial robustness and are almost better than FP networks. For example, $ACC$ of ResNet18 DoReFa 2-bit is 8.75\% lower than ResNet18 FP (see \tabref{tab:clean_acc}), while $WCAR$ of ResNet18 DoReFa 2-bit is 4.25\% higher than ResNet18 FP (see \tabref{tab:ADV_WCAR}). Moreover, \figref{fig:ResNet18_Adv} illustrates that under the same quantization method, the adversarial robustness of quantized ResNet18 models \textbf{increases} as the quantization bit-width \textbf{decreases}. These phenomena can also be observed in other network architectures, suggesting that quantization can provide defense against adversarial attacks to a certain extent.

(2) \emph{At the same quantization bit-width}, PACT outperforms other quantization methods under the worst-case adversarial robustness. For instance, RegNetX600M PACT 4-bit achieves a $WCAR$ of 6.21\%, while DoReFa and LSQ quantized models achieve $WCAR$ values of 2.74\% and 2.05\%, respectively (see \tabref{tab:ADV_WCAR}). However, when facing specific adversarial attacks, the robustness performance of quantization methods is not consistent across different model architectures (see \tabref{tab:ADV_AR_FGSMLinf} and \tabref{tab:ADV_AR_PGDLinf}). For ResNet18, ResNet50, and MobileNetV2 architectures, the LSQ quantization method demonstrates better robustness against FGSM-$\ell_{\infty}$ and PGD-$\ell_{\infty}$ attacks. However, for RegNetX600M architecture, the PACT method exhibits better robustness. 

(3) \emph{As for the model size and network architecture}, quantized models show similar trends to FP models. \emph{Regarding the model size}, we observe that quantized models with larger network capacity (\eg, FLOPs and Params) exhibit better adversarial robustness, consistent with the findings in \cite{tang2021robustart}. For instance, in the ResNet families, all quantized ResNet50 models demonstrate better robustness than ResNet18 against all adversarial attacks. \emph{Regarding the network architecture}, the adversarial robustness of quantized models follows the order: MobileNetV2 $>$ RegNetX600M $>$ ResNet, which coincides with the findings in \cite{tang2021robustart}. This highlights the significance of architecture design in achieving better adversarial robustness.

(4) \emph{As for the adversarial attack methods}, their attack capabilities in quantized models are generally consistent with those in FP models: AutoAttack $>$ PGD $>$ FGSM. However, quantized models demonstrate varying adversarial robustness against different attack methods. For instance, compared to DoReFa and PACT, LSQ performs better under FGSM and PGD attacks but is more vulnerable to AutoAttack.

\subsection{Evaluation of natural corruptions}

\begin{table*}[t] 
\centering

\caption{Natural Robustness of quantized models under all corruption methods. Results are shown in $mNR\textcolor{red}{\uparrow}(NCI\textcolor{green}{\downarrow})$. The best performers in each architecture are highlighted in bold.} 

\label{tab:mNR}

\begin{tabular}{@{}llccccc@{}}
\toprule
Model & Method & 2-bit & 4-bit & 6-bit & 8-bit & 32-bit \\ \midrule
\multirow{3}{*}{ResNet18} & DoReFa & 23.30 (62.61\%) & 30.88 (56.26\%) & 31.79 (55.32\%) & 31.70 (55.67\%) & \multirow{3}{*}{\pmb{32.78 (53.87\%)}} \\ 
 & PACT & NC & 30.36 (56.88\%) & 31.50 (55.74\%) & 31.69 (55.63\%) \\ 
 & LSQ & 26.42 (59.95\%) & 30.95 (56.11\%) & 31.67 (55.46\%) & 31.70 (55.55\%) \\  \midrule
\multirow{3}{*}{ResNet50} & DoReFa & NC & 38.87 (49.39\%) & 39.52 (48.75\%) & 39.74 (48.40\%) & \multirow{3}{*}{\pmb{40.37 (47.32\%)}} \\ 
 & PACT & NC & 38.53 (49.83\%) & 39.59 (48.60\%) & 39.60 (48.61\%) \\ 
 & LSQ & 29.64 (57.55\%) & 39.48 (48.72\%) & 40.04 (48.24\%) & 40.06 (48.34\%) \\  \midrule
\multirow{3}{*}{RegNetX600M} & DoReFa & NC & 33.58 (53.81\%) & 35.01 (52.55\%) & 35.23 (52.39\%) & \multirow{3}{*}{\pmb{35.48 (51.76\%)}} \\ 
 & PACT & NC & 32.17 (55.42\%) & 34.66 (53.04\%) & 35.31 (52.32\%) \\ 
 & LSQ & NC & 33.42 (53.93\%) & 34.72 (52.85\%) & 34.97 (52.60\%) \\  \midrule
\multirow{3}{*}{MobileNetV2} & DoReFa & NC & NC & NC & NC & \multirow{3}{*}{\pmb{33.19 (54.30\%)}} \\ 
 & PACT & NC & 29.70 (57.82\%) & 32.05 (55.81\%) & 32.40 (55.43\%) \\ 
 & LSQ & NC & 29.92 (57.45\%) & 31.72 (55.98\%) & 32.06 (55.72\%) \\  \bottomrule

\end{tabular}
\end{table*}

\begin{table*}[]
\centering
\caption{Natural robustness results for ResNet18 models are shown in $NR\textcolor{red}{\uparrow}$ for each corruption (\eg, Gauss). The most influential noise is marked in bold, and the least influential noise is underlined.}
\label{tab:NR_ResNet18}

\resizebox{\linewidth}{!}{%
\begin{tabular}{ll|c|ccc|cccc|cccc|cccc|cc@{}}
\toprule
   & \multicolumn{1}{l|}{} & \multicolumn{1}{c|}{} & \multicolumn{3}{c|}{Noise} & \multicolumn{4}{c|}{Blur} & \multicolumn{4}{c|}{Weather} & \multicolumn{4}{c|}{Digital} &  &  \\ \midrule
Quant. & \multicolumn{1}{l|}{Bit} & \multicolumn{1}{c|}{ACC} & Gauss & Shot & \multicolumn{1}{l|}{Impulse} & Defocus & Glass & Motion & \multicolumn{1}{l|}{Zoom} & Snow & Frost & Fog & Bright & Contrast & Elastic & Pixel & JPEG & mNR$\textcolor{red}{\uparrow}$ & NCI$\textcolor{green}{\downarrow}$ \\ \midrule

FP & 32 & 71.06 & 25.88 & 23.86 & \pmb{19.76} & 28.34 & 21.88 & 30.37 & 29.80 & 26.89 & 29.96 & 36.78 & \underline{ 60.05} & 34.32 & 38.50 & 39.59 & 45.77 & 32.78 & 53.87\% \\ \midrule

\multirow{4}{*}{DoReFa} & 2 & 62.31 & 15.38 & 13.55 & \pmb{9.26} & 17.28 & 16.49 & 21.40 & 21.11 & 18.44 & 22.03 & 25.25 & \underline{ 50.16} & 23.11 & 32.98 & 27.45 & 35.58 & 23.30 & 62.61\% \\
 & 4 & 70.60 & 25.23 & 22.78 & \pmb{16.94} & 24.68 & 20.70 & 27.95 & 27.05 & 25.24 & 29.34 & 33.91 & \underline{ 59.41} & 31.13 & 37.67 & 36.33 & 44.85 & 30.88 & 56.26\% \\
 & 6 & 71.15 & 26.24 & 23.69 & \pmb{18.70} & 25.89 & 20.78 & 28.73 & 27.68 & 25.86 & 29.79 & 35.41 & \underline{ 60.10} & 33.14 & 37.88 & 37.48 & 45.46 & 31.79 & 55.32\% \\
 & 8 & 71.51 & 25.68 & 22.89 & \pmb{17.88} & 25.96 & 20.70 & 28.77 & 27.72 & 25.88 & 29.81 & 35.57 & \underline{ 60.20} & 33.37 & 37.76 & 37.71 & 45.65 & 31.70 & 55.66\% \\ \midrule
\multirow{4}{*}{PACT} & 2 & NC & NC & NC & NC & NC & NC & NC & NC & NC & NC & NC &  NC & NC & NC & NC & NC & NC & NC \\
 & 4 & 70.41 & 24.88 & 22.46 & \pmb{17.10} & 23.96 & 20.49 & 27.36 & 26.70 & 24.58 & 28.66 & 33.33 & \underline{ 58.90} & 29.63 & 37.30 & 35.83 & 44.27 & 30.36 & 56.87\% \\
 & 6 & 71.17 & 25.68 & 23.06 & \pmb{17.98} & 25.54 & 20.77 & 28.56 & 27.54 & 25.56 & 29.58 & 34.91 & \underline{ 60.07} & 32.46 & 37.74 & 37.69 & 45.30 & 31.50 & 55.75\% \\
 & 8 & 71.43 & 25.56 & 22.89 & \pmb{17.95} & 26.07 & 20.68 & 28.72 & 27.66 & 25.85 & 29.77 & 35.54 & \underline{ 60.14} & 33.48 & 37.83 & 37.62 & 45.59 & 31.69 & 55.63\% \\ \midrule
\multirow{4}{*}{LSQ} & 2 & 65.97 & 19.85 & 17.79 & \pmb{12.70} & 20.79 & 18.62 & 24.45 & 23.84 & 20.36 & 24.49 & 28.54 & \underline{ 54.08} & 25.92 & 35.49 & 30.59 & 38.84 & 26.42 & 59.95\% \\
 & 4 & 70.52 & 25.79 & 23.08 & \pmb{17.54} & 25.01 & 20.79 & 28.04 & 27.12 & 25.36 & 29.37 & 33.78 & \underline{ 59.40} & 30.65 & 37.58 & 35.99 & 44.73 & 30.95 & 56.11\% \\
 & 6 & 71.10 & 26.31 & 23.67 & \pmb{18.37} & 25.78 & 21.10 & 28.84 & 27.98 & 26.12 & 29.87 & 34.77 & \underline{ 60.28} & 31.49 & 37.83 & 37.11 & 45.51 & 31.67 & 55.46\% \\
 & 8 & 71.31 & 26.21 & 23.45 & \pmb{18.34} & 25.91 & 21.03 & 28.91 & 28.06 & 26.09 & 29.74 & 34.86 & \underline{ 60.25} & 31.72 & 37.92 & 37.29 & 45.70 & 31.70 & 55.55\% \\ \bottomrule

\end{tabular}
}
\end{table*}

\begin{figure*}[ht]
  \includegraphics[width=1.0\linewidth]{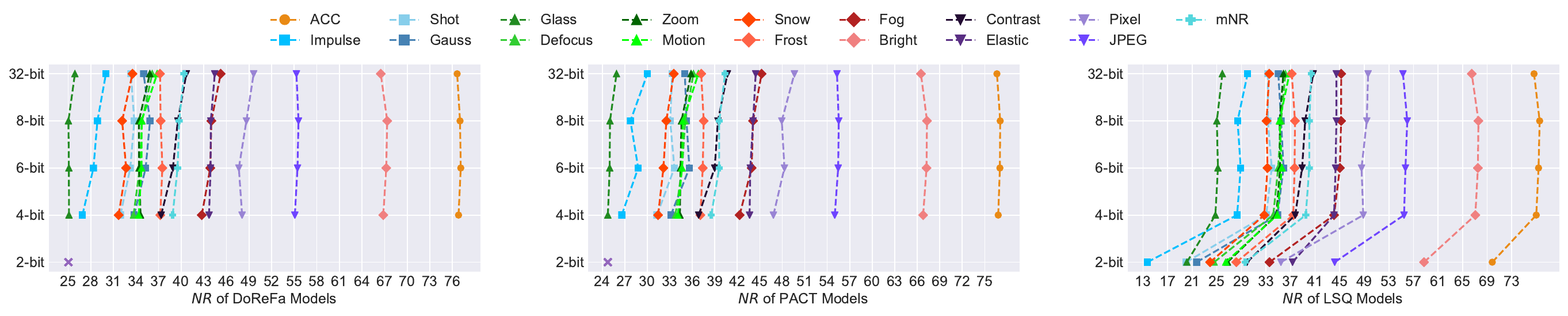}
  \caption{Natural robustness of ResNet50 models under specific corruption. From left to right: quantized by DoReFa, PACT, and LSQ respectively. The ``NC" models are labeled with `x' in the figure. }
  \label{fig:ResNet50_Avg}       
\end{figure*}

\begin{figure*}[ht]
  \includegraphics[width=1.0\linewidth]{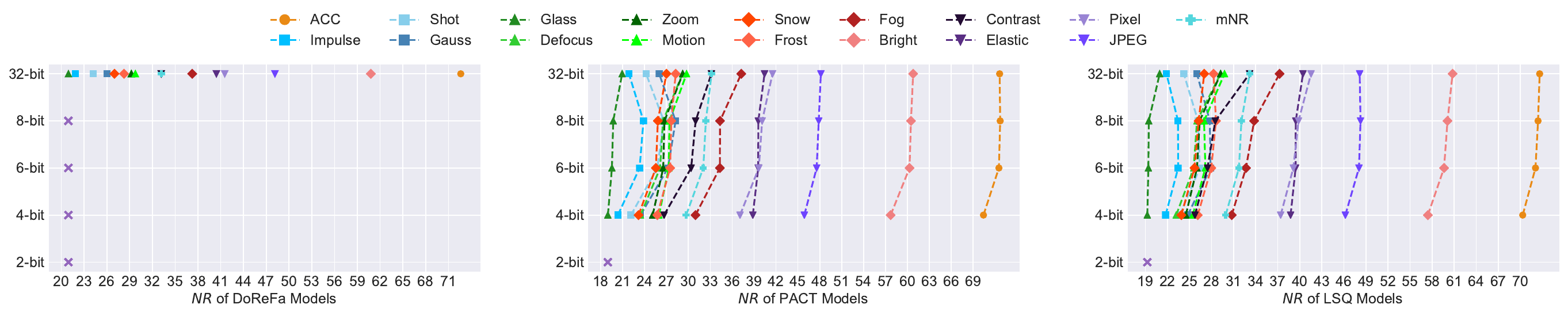}
  \caption{Natural robustness of MobileNetV2 models under specific corruption. From left to right: quantized by DoReFa, PACT, and LSQ respectively. The ``NC" models are labeled with `x' in the figure. }
  \label{fig:MobileNetV2_Avg}       
\end{figure*}

We first report the aggregated evaluation results (\ie, mean natural robustness) among fifteen corruption methods for all quantized models, shown in \tabref{tab:mNR}. Next, we explore the natural robustness results for different quantized models under each corruption method. For instance, we provide the natural robustness results for ResNet18, ResNet50, and MobileNetV2 architectures in \tabref{tab:NR_ResNet18}, \figref{fig:ResNet50_Avg}, and \figref{fig:MobileNetV2_Avg}, respectively. Due to the limit of space, more detailed results can be found on our website \cite{ourweb}. The natural robustness evaluation results for quantized models reveal several key observations as follows.

(1) \emph{Worse natural robustness vs FP models}. Despite achieving similar clean accuracy compared to the 32-bit model, quantized models are more susceptible to natural corruptions. Notably, the 2-bit quantized model is severely affected, with a decrease in performance that far exceeds the accuracy drop observed in clean accuracy. As an example, ResNet18 DoReFa 2-bit exhibits a $NCI$ of 62.61\%, which is higher than the corresponding FP model with 53.87\% $NCI$ (see \tabref{tab:mNR}), indicating that quantization can negatively impact the model's robustness against such perturbations. 
Furthermore, \figref{fig:ResNet50_Avg} demonstrates that under the same quantization method, the natural robustness of quantized ResNet50 models generally \textbf{increases} as the quantization bit-width \textbf{increases}. However, the rate of increase in robust performance gradually decreases.
Similar phenomena can also be observed in other network architectures, indicating that in real scenarios, using quantized models requires more attention than FP models. For instance, leveraging corruption data to enhance natural robustness becomes crucial in maintaining the performance of quantized models.

(2) \emph{At the same quantization bit-width}, quantization methods exhibit varied performance on different architectures. As shown in \tabref{tab:mNR}, on ResNet18 and ResNet50 architectures, LSQ achieves less relative performance drop (\ie, lower $NCI$ value) than DoReFa and PACT methods. However, on RegNetX600M and MobileNetV2 architectures, DoReFa and PACT may outperform other quantization methods across most bit-widths. When examining a specific architecture, \figref{fig:ResNet50_Avg} reveals a consistent relative trend among different quantization methods across various corruption methods (other architectures also show this trend). Specifically, the order of impact on model performance is as follows: \textbf{Noise} $>$ \textbf{Blur} $>$ \textbf{Weather} (except Brightness) $>$ \textbf{Digital}. This consistent trend emphasizes the varying degrees of sensitivity of quantized models to different types of corruption, and it remains independent of the specific quantization method used.

(3) \emph{Under the same network architecture}, network capacity could lead to better natural robustness.  Similar to the adversarial evaluation, quantized ResNet50 models here present higher $mNR$ and lower $NCI$ compared to ResNet18 under natural corruptions. \emph{While for different network architectures}, the natural robustness of quantized models follows the order:  ResNet $>$ RegNetX600M $>$ MobileNetV2, which is exactly opposite to the order observed under adversarial attacks.

(4) \emph{Regarding natural corruption methods}, our conclusions are as follows. For ResNet18, ResNet50, and RegNetX600M, \textbf{impulse noise} has the most severe impact on the model's robustness. For MobileNetV2, \textbf{glass blur} is the most harmful corruption to the model's robustness. Among all the corruption methods, \textbf{brightness} has the least impact on the model's robustness. As shown in \tabref{tab:NR_ResNet18}, ResNet18 quantized models exhibit an average decrease of about 76.32\% under impulse noise, whereas the average decrease is only about 16.27\% under brightness weather corruption. Similarly, MobileNetV2 quantized models show an average decrease of about 72.96\% under glass blur, while the average decrease is only about 17.39\% under brightness.

\begin{table}[ht] 
\centering

\caption{Natural Robustness of quantized models under all corruption sequences. Results are shown in $mFP\textcolor{green}{\downarrow}$. The best performers in each architecture are highlighted in bold.}

\label{tab:mFP}

\resizebox{\columnwidth}{!}{%
\begin{tabular}{@{}llccccc@{}}
\toprule
Model & Method & 2-bit & 4-bit & 6-bit & 8-bit & 32-bit \\ \midrule
\multirow{3}{*}{ResNet18} & DoReFa & 0.283 & 0.152 & 0.113 & 0.106 & \multirow{3}{*}{\pmb{0.098}} \\
 & PACT & NC & 0.182 & 0.133 & 0.106 &  \\
 & LSQ & 0.245 & 0.143 & 0.113 & 0.106 &  \\ \midrule
\multirow{3}{*}{ResNet50} & DoReFa & NC & 0.124 & 0.088 & 0.082 & \multirow{3}{*}{\pmb{0.072}} \\
 & PACT & NC & 0.139 & 0.101 & 0.082 &  \\
 & LSQ & 0.207 & 0.111 & 0.086 & 0.079 &  \\ \midrule
\multirow{3}{*}{RegNetX600M} & DoReFa & NC & 0.163 & 0.114 & 0.106 & \multirow{3}{*}{\pmb{0.097}} \\
 & PACT & NC & 0.201 & 0.137 & 0.105 &  \\
 & LSQ & NC & 0.158 & 0.120 & 0.108 &  \\ \midrule
\multirow{3}{*}{MobileNetV2} & DoReFa & NC & NC & NC & NC & \multirow{3}{*}{\pmb{0.098}} \\
 & PACT & NC & 0.198 & 0.119 & 0.103 &  \\
 & LSQ & NC & 0.187 & 0.140 & 0.119 &  \\ \bottomrule
\end{tabular}
}

\end{table}

\begin{table*}[ht]
\centering
\caption{Natural Robustness of ResNet18 under corruption sequences. Results for each corruption (\eg, Gauss) are shown in $FP\textcolor{green}{\downarrow}$. The most
influential noise is marked in bold, and the least influential noise is underlined.}
\label{tab:FP_ResNet18}
\begin{tabular}{@{}ll|cc|cc|cc|cccc|c@{}}
\toprule
   & \multicolumn{1}{l|}{}  & \multicolumn{2}{c|}{Noise} & \multicolumn{2}{c|}{Blur} & \multicolumn{2}{c|}{Weather} & \multicolumn{4}{c|}{Digital} &    \\ \midrule
  Quant. & \multicolumn{1}{l|}{Bit}  & Gauss &  \multicolumn{1}{l|}{Shot}  & Motion & \multicolumn{1}{l|}{Zoom} & Snow  & Bright & Translate & Rotate & Tilt & Scale & $mFP\textcolor{green}{\downarrow}$\\ \midrule

FP & 32 & 0.164 & \pmb{0.207} & 0.076 & 0.049 & 0.091 & \underline{0.039} & 0.068 & 0.090 & 0.050 & 0.145 & 0.098 \\ \midrule

\multirow{4}{*}{DoReFa} & 2 & 0.293 & \pmb{0.335} & 0.315 & 0.318 & 0.335 & \underline{0.233} & 0.223 & 0.254 & 0.230 & 0.298 & 0.283 \\
 & 4 & 0.185 & \pmb{0.223} & 0.156 & 0.151 & 0.172 & \underline{0.105} & 0.110 & 0.133 & 0.106 & 0.178 & 0.152 \\
 & 6 & 0.170 & \pmb{0.210} & 0.099 & 0.078 & 0.112 & \underline{0.056} & 0.080 & 0.102 & 0.065 & 0.154 & 0.113 \\
 & 8 & 0.169 & \pmb{0.212} & 0.087 & 0.061 & 0.102 & \underline{0.046} & 0.075 & 0.097 & 0.057 & 0.151 & 0.106 \\ \midrule
\multirow{4}{*}{PACT} & 2 & NC & NC & NC & NC & NC & NC & NC & NC & NC & NC & NC \\
 & 4 & 0.199 & \pmb{0.235} & 0.201 & 0.201 & 0.215 & \underline{0.139} & 0.135 & 0.158 & 0.137 & 0.198 & 0.182 \\
 & 6 & 0.176 & \pmb{0.217} & 0.130 & 0.120 & 0.142 & \underline{0.083} & 0.094 & 0.116 & 0.086 & 0.163 & 0.133 \\
 & 8 & 0.169 & \pmb{0.210} & 0.088 & 0.061 & 0.102 & \underline{0.046} & 0.075 & 0.097 & 0.057 & 0.152 & 0.106  \\ \midrule
\multirow{4}{*}{LSQ} & 2 & 0.254 & \pmb{0.291} & 0.275 & 0.277 & 0.295 & \underline{0.202} & 0.188 & 0.218 & 0.196 & 0.259 & 0.245 \\
 & 4 & 0.175 & \pmb{0.212} & 0.148 & 0.140 & 0.159 & \underline{0.102} & 0.104 & 0.126 & 0.099 & 0.170 & 0.143 \\
 & 6 & 0.163 & \pmb{0.202} & 0.103 & 0.086 & 0.112 & \underline{0.067} & 0.080 & 0.102 & 0.068 & 0.151 & 0.113 \\
 & 8 & 0.160 & \pmb{0.200} & 0.092 & 0.070 & 0.102 & \underline{0.059} & 0.076 & 0.097 & 0.061 & 0.147 & 0.106 \\ \bottomrule 

\end{tabular}
\end{table*}

\begin{figure*}[ht]
  \includegraphics[width=1.0\linewidth]{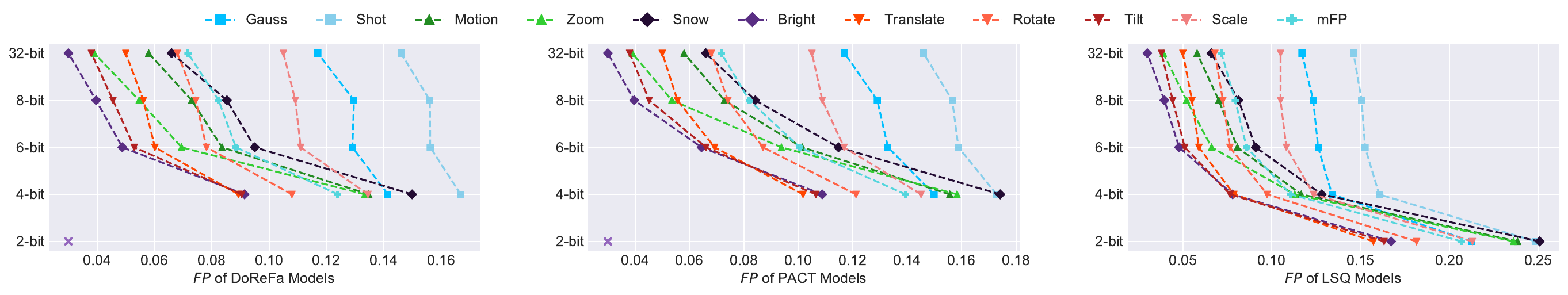}
  \caption{Natural robustness of ResNet50 models under corruption sequences. From left to right: quantized by DoReFa, PACT, and LSQ respectively. The ``NC" models are labeled with `x' in the figure. }
  \label{fig:ResNet50_FP}       
\end{figure*}

\begin{figure*}[ht]
  \includegraphics[width=1.0\linewidth]{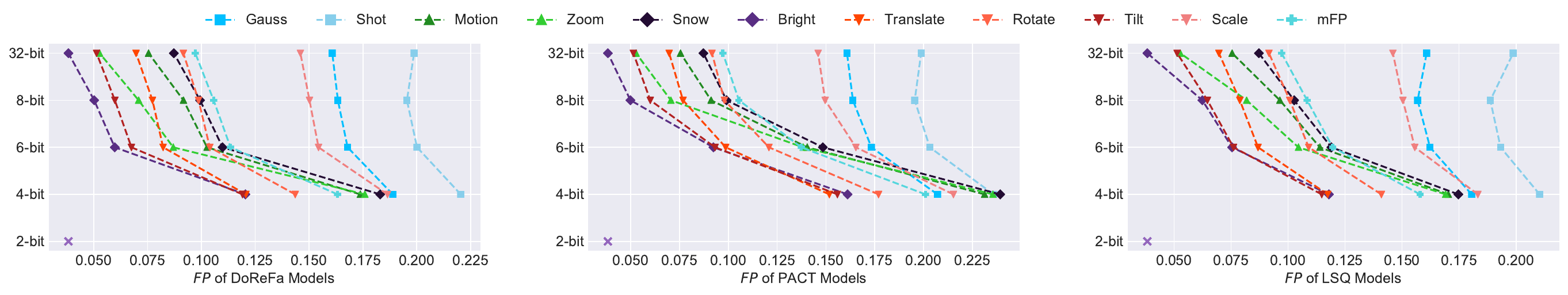}
  \caption{Natural robustness of RegNetX600M models under corruption sequences. From left to right: quantized by DoReFa, PACT, and LSQ respectively. The ``NC" models are labeled with `x' in the figure. }
  \label{fig:RegNetX600M_FP}       
\end{figure*}

For natural robustness under corruption sequences, we report the evaluation results in \tabref{tab:mFP}, \tabref{tab:FP_ResNet18}, \figref{fig:ResNet50_FP}, and \figref{fig:RegNetX600M_FP}. \tabref{tab:mFP} shows the mean Flip Probability of quantized models against multiple sequences. \tabref{tab:FP_ResNet18}, \figref{fig:ResNet50_FP}, and \figref{fig:RegNetX600M_FP} 
reports the detailed robustness results under specific corruption sequences for ResNet18, ResNet50 and RegNetX600M, respectively. Other results can be found on our website \cite{ourweb}. From these results, we can draw several conclusions about natural robustness under dynamic and continuous corruptions, which mostly align with our observations from static natural corruption evaluations.

(1) \emph{Worse dynamic natural robustness vs FP models}. Coinciding with the static natural corruptions, quantized models also exhibit inferior performance compared to FP models under dynamic corruption sequences. Furthermore, the 2-bit quantized models demonstrate extreme instability. For example, ResNet18 DoReFa 2-bit exhibits a $mFP$ of 0.283, which is nearly three times higher than that of the ResNet18 FP model (0.098 $mFP$). As depicted in \figref{fig:ResNet50_FP}, we can find that the natural robustness increases with increasing bit-width (from right to left in each subfigure).

(2) \emph{At the same quantization bit-width}, quantization methods under dynamic natural corruptions also exhibit similar trends as observed under static natural corruptions. For example, LSQ achieves better robustness on the ResNet architecture but may not perform as dominantly on lightweight architectures such as RegNetX600M and MobileNetV2.

(3) \emph{As for specific corruption sequences}, shot noise has the most severe impact on all models, while brightness remains the least harmful. Additionally, we observe that for RegNetX600M models, the FP model is not the most robust under shot noise (see \figref{fig:RegNetX600M_FP}), which is inconsistent with the behavior observed in ResNet models (see \figref{fig:ResNet50_FP}).

\subsection{Evaluation of Systematic Noises}

\begin{table*}[ht] 
\centering
\label{tab:mean_sys}

\caption{Mean Systematic Robustness of models under all noises. Results are shown in $ACC_s\textcolor{red}{\uparrow}(SNI\textcolor{green}{\downarrow})\pm SR \textcolor{green}{\downarrow} $.}
\resizebox{\linewidth}{!}{%
\begin{tabular}{@{}llrrrrr@{}}
\toprule
Model & Method & 2-bit & 4-bit & 6-bit & 8-bit & 32-bit \\ \midrule
\multirow{3}{*}{ResNet18} & DoReFa & 59.51 (4.49\%)$\pm$2.32 & 68.67 (2.73\%)$\pm$1.83  & 69.43 (2.42\%)$\pm$1.69 & 69.43 (2.91\%)$\pm$1.95 & \pmb{\multirow{3}{*}{70.31 (1.06\%)$\pm$0.53}} \\ 
 & PACT & NC & 68.38 (2.88\%)$\pm$1.85 & 69.25 (2.70\%)$\pm$1.84 & 69.38 (2.87\%)$\pm$1.98\\ 
 & LSQ & 63.52 (3.71\%)$\pm$2.05 & 68.47 (2.91\%)$\pm$1.90 & 69.16 (2.73\%)$\pm$1.87 & 69.43 (2.64\%)$\pm$1.82 \\  \midrule
\multirow{3}{*}{ResNet50} & DoReFa & NC & 75.36 (1.89\%)$\pm$1.69  & 75.62 (1.93\%)$\pm$1.67 & 75.42 (2.06\%)$\pm$1.70  & \pmb{\multirow{3}{*}{76.08 (0.72\%)$\pm$0.49}} \\ 
 & PACT & NC & 75.32 (1.93\%)$\pm$1.71  & 75.61 (1.83\%)$\pm$1.67  & 75.51 (2.01\%)$\pm$1.76  \\ 
 & LSQ & 67.72 (3.02\%)$\pm$2.18 & 75.54 (1.88\%)$\pm$1.60 & 76.07 (1.67\%)$\pm$1.62 & 76.06 (1.91\%)$\pm$1.69  \\  \midrule
\multirow{3}{*}{RegNetX600M} & DoReFa & NC & 70.92 (2.45\%)$\pm$1.51   & 72.15 (2.22\%)$\pm$1.51& 72.49 (2.04\%)$\pm$1.48 & \pmb{\multirow{3}{*}{72.92 (0.86\%)$\pm$0.57}} \\ 
 & PACT & NC & 70.27 (2.62\%)$\pm$1.54  & 72.21 (2.15\%)$\pm$1.60  & 72.49 (2.11\%)$\pm$1.46 \\ 
 & LSQ & NC & 70.79 (2.41\%)$\pm$1.48  & 72.01 (2.21\%)$\pm$1.51 & 72.20 (2.13\%)$\pm$1.47  \\  \midrule
\multirow{3}{*}{MobileNetV2} & DoReFa & NC & NC & NC & NC & \pmb{\multirow{3}{*}{71.71 (1.25\%)$\pm$0.52 }} \\ 
 & PACT & NC & 68.30 (3.00\%)$\pm$1.88   & 70.52 (2.76\%)$\pm$1.71 & 70.75 (2.68\%)$\pm$1.70\\ 
 & LSQ & NC & 68.15 (3.09\%)$\pm$1.98  & 70.01 (2.83\%)$\pm$1.92& 70.41 (2.75\%)$\pm$1.78 \\  \bottomrule
\end{tabular}
}
\end{table*}

\begin{table*}[!ht]
\label{tab:sys_ResNet18}
\caption{Systematic Robustness of ResNet18 models under each systematic noise. Results for each noise (\eg, Bilinear) are shown in $ACC_s\textcolor{red}{\uparrow}$. The most
influential noise is marked in bold, and the least influential noise is underlined.}
\resizebox{\linewidth}{!}{%
\begin{tabular}{@{}ll|c|cccccc|ccccc|ccc|cc@{}}
\toprule
& \multicolumn{1}{l|}{} & \multicolumn{1}{c|}{} & \multicolumn{6}{c|}{Resize mode in Pillow} & \multicolumn{5}{c|}{Resize mode in OpenCV} & \multicolumn{3}{c|}{Decoder}  &  &  \\ \midrule
 Quant. & \multicolumn{1}{l|}{Bit} & \multicolumn{1}{c|}{ACC} & Bilinear & Nearest & Box & Hamming & Cubic & Lanczos & Bilinear & Nearest & Area & Cubic & Lanczos & Pillow & OpenCV & FFmpeg  & Mean$\textcolor{red}{\uparrow}$ & $SR\textcolor{green}{\downarrow}$ \\ \midrule

 FP & 32 & 71.06 & 70.21 & 69.15 & 70.64 & 70.73 & 70.66 & 70.78 & 70.82 & 69.21 & 70.71 & 70.53 & 70.21 & 70.21 & 70.22 & 70.22 & 70.31 & 0.53 \\ \midrule

 \multirow{4}{*}{LSQ} & 2 & 65.97 & 65.01 & \pmb{59.48} & 63.74 & 64.79 & \underline{65.04} & 64.48 & 63.58 & \pmb{59.44} & \underline{64.95} & 62.15 & 61.56 & 65.01 & 65.01 & 65.00 & 63.52 & 2.05 \\ 
 & 4 & 70.52 & \underline{69.82} & \pmb{64.79} & 68.85 & 69.57 & 69.77 & 69.46 & 68.71 & \pmb{64.74} & \underline{69.76} & 67.01 & 66.51 & 69.82 & 69.85 & 69.87 & 68.47 & 1.90 \\
 & 6 & 71.10 & \underline{70.51} & \pmb{65.49} & 69.46 & 70.37 & 70.48 & 70.18 & 69.36 & \pmb{65.78} & \underline{70.44} & 67.69 & 66.91 & 70.51 & 70.51 & 70.50 & 69.16 & 1.87 \\
 & 8 & 71.31 & \underline{70.73} & \pmb{66.07} & 69.77 & 70.64 & \underline{70.73} & 70.42 & 69.67 & \pmb{66.05} & \underline{70.73} & 67.95 & 67.09 & 70.73 & 70.72 & 70.68 & 69.43 & 1.82 \\ \midrule
\multirow{4}{*}{DoReFa} & 2 & 62.31 & \underline{61.14} & \pmb{54.77} & 59.86 & 60.98 & 60.98 & 60.71 & 59.79 & \pmb{55.04} & \underline{61.22} & 57.88 & 57.26 & 61.14 & 61.07 & 61.24 & 59.50 & 2.32 \\
 & 4 & 70.60 & 69.88 & \pmb{65.03} & 69.03 & 69.75 & \underline{70.00} & 69.65 & 69.01 & \pmb{64.97} & \underline{69.94} & 67.45 & 66.94 & 69.88 & 69.94 & 70.00 & 68.68 & 1.83 \\
 & 6 & 71.15 & 70.61 & \pmb{66.08} & 69.67 & 70.48 & \underline{70.66} & 70.35 & 69.67 & \pmb{66.14} & \underline{70.55} & 68.26 & 67.62 & 70.61 & 70.61 & 70.70 & 69.43 & 1.69 \\
 & 8 & 71.51 & \underline{70.88} & \pmb{65.88} & 69.72 & 70.68 & 70.72 & 70.47 & 69.67 & \pmb{65.86} & \underline{70.75} & 67.84 & 66.86 & 70.88 & 70.87 & 70.97 & 69.43 & 1.95 \\ \midrule
\multirow{4}{*}{PACT} & 2 & NC & NC & NC & NC & NC & NC & NC & NC & NC & NC & NC & NC & NC & NC & NC & NC & NC \\
 & 4 & 70.41 & 69.69 & \pmb{64.76} & 68.71 & 69.42 & \underline{69.71} & 69.31 & 68.59 & \pmb{64.71} & \underline{69.64} & 67.19 & 66.39 & 69.69 & 69.60 & 69.88 & 68.38 & 1.85 \\
 & 6 & 71.17 & \underline{70.62} & \pmb{65.66} & 69.40 & 70.39 & 70.51 & 70.14 & 69.47 & \pmb{65.91} & \underline{70.58} & 67.76 & 67.12 & 70.62 & 70.67 & 70.63 & 69.25 & 1.84 \\
 & 8 & 71.43 & \underline{70.89} & \pmb{65.71} & 69.65 & 70.67 & 70.68 & 70.39 & 69.72 & \pmb{65.78} & \underline{70.71} & 67.74 & 66.75 & 70.89 & 70.84 & 70.84 & 69.38 & 1.98 \\
\bottomrule
\end{tabular}
}
\end{table*}

\begin{figure*}[ht]
  \includegraphics[width=1.0\linewidth]{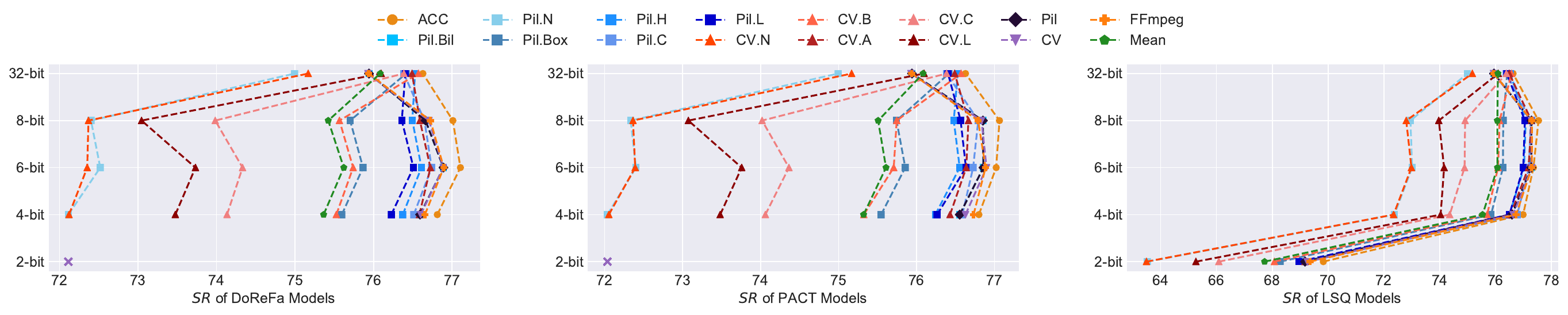}
  \caption{Systematic robustness of ResNet50 models under specific noises. From left to right: quantized by DoReFa, PACT, and LSQ respectively. The ``NC" models are labeled with `x' in the figure. }
  \label{fig:ResNet50_Sys}       
\end{figure*}

\begin{figure*}[ht]
  \includegraphics[width=1.0\linewidth]{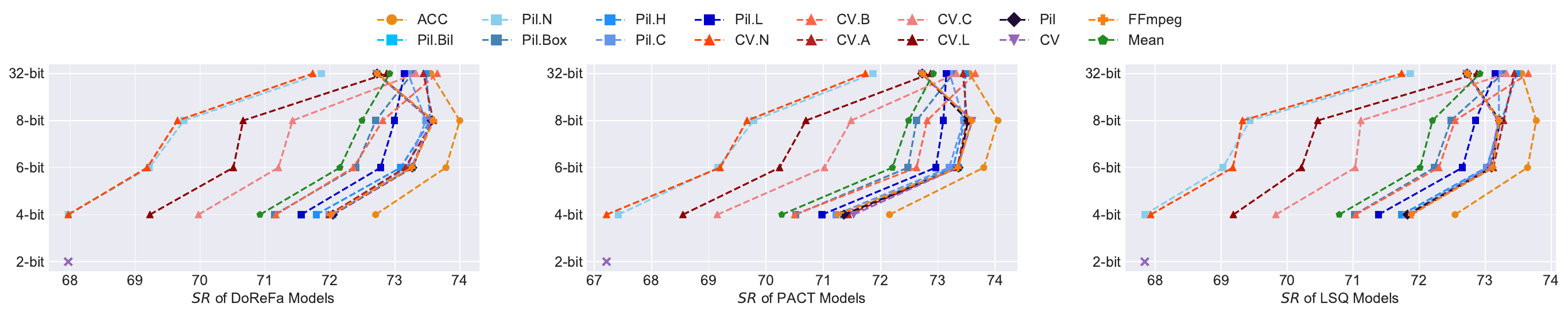}
  \caption{Systematic robustness of RegNetX600M models under specific noises. From left to right: quantized by DoReFa, PACT, and LSQ respectively. The ``NC" models are labeled with `x' in the figure. }
  \label{fig:RegNetX600M_Sys}       
\end{figure*}

We report the aggregated evaluation results of systematic robustness among fourteen systematic noises for all quantized models, shown in \tabref{tab:mean_sys}. Then, we explore the performance for different quantized models under specific systematic noise. For instance, we provide the systematic robustness results for ResNet18, ResNet50, and RegNetX600M architectures in \tabref{tab:sys_ResNet18}, \figref{fig:ResNet50_Sys}, and \figref{fig:RegNetX600M_Sys}, respectively. Due to the limit of space, more detailed results can be found on our website \cite{ourweb}. The evaluation results for quantized models provide valuable insights into their systematic robustness. Several key observations are as follows:

(1) \emph{Worse systematic robustness vs FP models}. 
Despite the prevalent system noise in deployed environments not causing a significant drop in the prediction accuracy of quantized models, their performance in various deployment scenarios exhibits considerable instability. In contrast, FP models demonstrate little fluctuation. For instance, on the ResNet18 architecture, all quantized models exhibit at least 3.43 times higher instability ($SR$) than the FP ResNet18, confirming the sensitivity of quantized models to system noise. In particular, 2-bit quantized models experience significant impacts under systematic noise (\eg, ResNet18 DoReFa 2-bit shows a drop of 4.49\% in $ACC_s$ and an increase of 4.38 times in instability). As for the same quantization method, \textbf{lower-bit}
 models present \textbf{less robustness} (\ie, lower stability) generally. It indicates that maintaining consistency between the deployment and training process is crucial to avoid unnecessary accuracy loss.

(2) \emph{At the same quantization bit-width}, Quantization methods show varied performance across different architectures. Generally, LSQ shows dominance in ResNet and RegNetX600M architectures, while PACT performs better in MobileNetV2. When focusing on a specific architecture (such as ResNet50 in \figref{fig:ResNet50_Sys} and RegNetX600M in \figref{fig:RegNetX600M_Sys}), we can observe the similar relative impact across different decoders and resize modes under three quantization methods.  

(3) \emph{Under the same network architecture}, network capacity could lead to better systematic robustness. Quantized ResNet50 models present lower $SR$ compared to ResNet18 under systematic noises. \emph{While for different network architectures}, the systematic robustness of quantized models follows the order:  RegNetX600M $>$ ResNet $>$ MobileNetV2. It is worth noting that the RegNetX600M FP model exhibits the worst systematic robustness, while surprisingly, the quantized RegNetX600M models demonstrate the best systematic robustness among all architectures.

(4) \emph{When considering specific systematic noise}, we draw the following conclusions. Among 14 systematic noises, the \textbf{nearest neighbor interpolation} methods in Pillow and OpenCV libraries have the most harmful impact on the model performance, which induce nearly a 6\% decrease in performance for the 2-bit ResNet18 models (see \tabref{tab:sys_ResNet18}). By contrast, the least impactful noises on model performance are \textbf{bilinear interpolation} and \textbf{cubic interpolation} in the Pillow library, as well as \textbf{area interpolation} in the OpenCV library. For example, under bilinear interpolation of Pillow library, 2-bit ResNet18 models only present 1.67\% decrease in performance on average. Regarding the different decoders, the performance of quantized models has only minor fluctuations among the three.
\section{Conclusion}
This paper presents a benchmark named RobustMQ, aiming to evaluate the robustness of quantized models under various perturbations, including adversarial attacks, natural corruptions, and systematic noises. The benchmark evaluates four classical architectures and three popular quantization methods with four different bit widths. 
The comprehensive results empirically provide valuable insights into the performance of quantized models in various scenarios. Some of the key observations are as follows. (1)
Quantized models exhibit higher adversarial robustness than their floating-point counterparts, but are more vulnerable to natural corruptions and systematic noises. (2) Under the same quantization method, we observe that as the quantization bit-width increases, the adversarial robustness decreases, the natural robustness increases, and the systematic robustness increases. (3) Among the 15 corruption methods, \textit{impulse noise} consistently exhibits the most harmful impact on ResNet18, ResNet50, and RegNetX600M models, while \textit{glass blur} is the most harmful corruption on MobileNetV2 models. On the other hand, \textit{brightness} is observed to be the least harmful corruption for all models. (4) Among the 14 systematic noises, the \textit{nearest neighbor interpolation} has the highest impact, while bilinear interpolation, cubic interpolation in Pillow, and area interpolation in OpenCV are the three least harmful
ones across different architectures. We hope that our benchmark will significantly contribute to the assessment of the robustness of quantized models. And the insights gained from our study could further support the development and deployment of robust quantized models in real-world scenarios.

\begin{small}
\vspace{.3in} \noindent \textbf{Data Availability:}
The datasets generated during and analyzed during the current study 
are available on our website \cite{ourweb}.
%

\vspace{.3in} \noindent \textbf{Competing Interests:}
The authors declare no competing interests.

\vspace{.3in} \noindent \textbf{Authors’ Contributions:} The first draft of the manuscript was written by YX and AL. YX and TZ collected data and conducted the experiment. 
YX, AL, HQ, and JG contributed to the analysis and manuscript preparation, and XL finalized the manuscript. 
All authors commented on previous versions of the manuscript and have read and approved the final manuscript. AL and XL conceived the original concept and supervised the study.


\end{small}

\begin{acknowledgements}
This work was supported by the National Key R\&D Program of China (2022ZD0116310), the National Natural Science Foundation of China (62022009 and 62206009), and the State Key Laboratory of Software Development Environment.

\end{acknowledgements}

\bibliographystyle{unsrt}
\bibliography{reference}

\end{document}